# Tree Index: A New Cluster Evaluation Technique

A. H. Beg
*School of Computing and Mathematics*
*Charles Sturt University*
*Panorama Avenue, Bathurst 2795, Australia*
E-mail: abeg@csu.edu.au

Md Zahidul Islam
*School of Computing and Mathematics*
*Charles Sturt University*
*Panorama Avenue, Bathurst 2795, Australia*
E-mail: zislam@csu.edu.au

Vladimir Estivill-Castro
*Departament de Tecnologies de la Informació i les Comunicacions,*
*Universitat Pompeu Fabra, Roc Boronat, 138, 08018 Barcelona, Spain*
E-mail: vladimir.estivill@upf.edu

**Abstract**—We introduce a cluster evaluation technique called Tree Index. Our Tree Index algorithm aims at describing the structural information of the clustering rather than the quantitative format of cluster-quality indexes (where the representation power of a clustering is some cumulative error similar to vector quantization). Our Tree Index is finding margins amongst clusters for easy learning without the complications of Minimum Description Length. Our Tree Index produces a decision tree from the clustered data set, using the cluster identifiers as labels. It combines the entropy of each leaf with their depth. Intuitively, a shorter tree with pure leafs generalizes the data well (the clusters are easy to learn because they are well separated). So, the labels are meaningful clusters. If the clustering algorithm does not separate well, trees learned from their results will be large and too detailed. We show that, on the clustering results (obtained by various techniques) on a brain dataset, Tree Index discriminates between reasonable and non-sensible clusters. We confirm the effectiveness of Tree Index through graphical visualizations. Tree Index evaluates the sensible solutions higher than the non-sensible solutions while existing cluster-quality indexes fail to do so.

## 1. INTRODUCTION

### A. Clustering evaluation techniques

Clustering is a well-known and an important technique in the area of data mining and data science. The objective of a clustering technique is to group similar records in a cluster and dissimilar records in different groups [1-6]. Clustering has enhanced data analysis in very many areas such as business [19], machine learning [20] social network analysis [11] and medical imaging [22].

Measuring the quality of a clustering solution requires an evaluation technique. There are two types of clustering-quality evaluation methods: internal clustering evaluation techniques and external clustering evaluation techniques. The external evaluation techniques evaluate clustering quality based on the external information such as pre-existing class labels. There are some well-known external evaluation techniques such as F-measure [1, 5, 6], Entropy [1] and Purity [1].

The internal cluster evaluation techniques evaluate the goodness of a cluster without class labels. They are known as cluster-quality indexes, and standard indexes are Davis-Bouldin (DB) Index [7], Silhouette Coefficient [1, 8], Sum of Square Error (SSE) [1] and COSEC [2]. These cluster-quality indexes are commonly used as the fitness function of genetic algorithms for clustering. Each index or coefficient is typically a formulation of different inductive criteria in a similar way as clustering algorithm can be seen as aiming to optimize a particular inductive principle [9]. For example, GenClust [2] uses COSEC as its fitness function to evaluate the quality of a chromosome. Generally, COSEC produces high values for a chromosome having a high number of clusters. HeMI [10] uses DB Index as the fitness function where a chromosome with a low number of clusters typically obtains a better DB value. GAGR [4] uses SSE as the fitness function where a chromosome with a high number of clusters tends to obtain a better SSE value. Hence, the standard cluster-quality indexes are biased towards clustering results having either too many or too few groups.

### B. Limitations of some existing standard cluster-quality indexes

To explore the clustering-quality evaluation methods and the quality of the existing clustering techniques we use a brain dataset (CHB-MIT Scalp) [13, 14] (see Section 3.4) as an example which is available from https://archive.physionet.org/cgi-bin/atm/ATM. We plot the dataset so that we can graphically visualise the clusters (see Fig. 1). Because of its origin, we know that this dataset has two types of records: seizure and non-seizure. We can also see in Fig. 1 that there are two clusters of records. We then apply the existing clustering techniques on this dataset and plot their clustering results.

We find that some recent and state of the art clustering techniques such as GAGR [4], GenClust [2] and HeMI [10] do not produce reasonable clusters. GenClust produces 447 clusters (see Fig. 2) which is not reasonable as the number of clusters for this dataset is only two. GAGR produces 56 clusters (see Fig. 3). HeMI produces two clusters (see Fig. 4) where one cluster contains one record, and the other cluster contains all remaining records.

We also evaluate the clustering quality of the existing techniques based on the internal and external evaluation criteria (see Table 1). While those clustering techniques we evaluated produce non-sensible clustering results, they achieve better quality values (compared to a reasonable clustering solution) based on the existing evaluation criteria. Therefore, a robust cluster validation method is also highly required to evaluate sensible and non-sensible clustering solutions.

## C. Novel components of the proposed technique and their advantages

In this paper, we propose the evaluation technique Tree Index. Our new evaluation technique Tree Index is capable of discriminating between a reasonable and a non-sensible clustering solution. Through our empirical analysis across existing clustering evaluation techniques (see Section 2.1) we observe that the existing clustering-quality evaluation techniques produce counter-intuitive evaluation values. Sometimes they produce higher evaluation values for non-sensible clustering solutions but lower values for reasonable clustering solutions. Sometimes they produce higher evaluation values for both: the reasonable and non-sensible clustering solutions which is not useful for assessing clustering quality.

Therefore, Tree Index uses as labels the classification proposed by a clustering algorithm to produce a decision tree [11, 12]. But we then calculate the entropy [1] for each leaf (i.e. the entropy of the distribution of class values within the leaf) and learn the depth of the leaf in the tree.

Based on the entropy and depth of a leaf, for all leaves, we then compute a clustering quality. The underlying idea here is that a good clustering result should create labels for the record that should be easy to learn [17]. Therefore, the decision trees built from such clustering results should not be too complex, not showing paths that are too deep. Moreover, good clustering results should lead to a decision tree having homogeneous leaves (i.e. low entropy). On the other hand, if the clustering result is bad (hard to learn), then we expect the resulting tree will struggle to find boundaries among the classes, an aspect that which will be reflected by heterogeneous leaves (i.e. high entropy) with high overall depth. Imagine an extreme example where the labels are assigned completely randomly (i.e. a very bad quality clustering), then it will be almost impossible for a tree to find any suitable pattern and the leaves are likely to be very heterogeneous and the tree will create many tests to separate them, resulting in deep branches.

## D. Main contributions of this study

In this paper we propose a new cluster evaluation technique called **Tree Index**. The Contributions in **Tree Index** is as follows.

- We propose a novel cluster quality criterion called Tree Index.
- We graphically plot various clustering solutions. Based on the visualisation of clustering, we then validate the correctness of the Tree Index evaluation. See Section 2 (B), Section 3(C) and Section 3(D).
- We also demonstrate the limitations of existing evaluation techniques by comparing their evaluation results with the cluster visualisation.

The rest of the paper is organized as follows: The proposed technique is described in Section 2. In Section 3, we discuss experimental results and Section 4 provides the concluding remarks.

## 2. OUR TECHNIQUE

### A. Basic concepts of our clustering evaluation technique Tree Index

We now discuss the basic concepts behind our proposed cluster evaluation technique, **Tree Index.**

We first contrast some reasonable with some non-sensible clustering solutions, despite that standard cluster-quality indexes produce opposite quality assessments. We use a brain data set called CHB-MIT Scalp EEG [13, 14] (see Section 3.4). Fig.1 shows the two-cluster structure, seizure and non-seizure, that domain experts provide. The original data set [13, 14] contains 9 non-class attributes and the class attribute with the two possible values: seizure and non-seizure. Fig.1 only display three non-class attributes (named `Max`, `Min` and `Std`). The dots shape represents the non-seizure class and the plus signs represents the seizure records. From Fig.1, we can see the existence of two clusters; one for mainly the seizure records and the other one for mainly non-seizure records.

Fig. 2 shows a non-sensible clustering result which is produced by GenClust on the brain data set. It appears to be a non-sensible clustering solution since it produces 477 clusters where the known number of clusters of this data set is only two. Fig. 2 uses 477 different shapes such as dots, plus sign, and triangles for each cluster. Note that the clustering algorithm uses only the non-class attributes.

Fig.3 shows another non-sensible clustering result which is obtained by GAGR on the brain data set. GAGR generates 56 clusters, which is also not reasonable.

Fig. 4 shows another non-sensible clustering result which is obtained by HeMI on the same brain data set. HeMI generates two clusters but the number of records in one cluster is one and all other records belong to the other cluster, this grouping is also not reasonable. However, it also achieves good cluster quality values for quality indexes such as F-measure, Purity, Silhouette Coefficient, XB Index and DB Index (refer to Table 1).

Table 1, when there is a clear winner, we have shaded the cells representing the highest evaluation value for an index (column) among the clustering techniques (the rows).

Fig. 5 shows an example of a reasonable clustering solution where it produces two clusters. Cluster 1 contains mostly the non-seizure records and Cluster 2 contains mostly the seizure records. We plot Fig. 5 based on the original orientation of records in the data set.

### B. Tree Index

From our empirical analysis, we observe that directly using a numerical cluster-quality index fails to produce distinctive optimization values to discriminate reasonable clustering solutions from non-sensible clustering solutions (see Table 1). As the number $n$ of data items and the dimension $d$ increases, these cluster-quality indexes will suffer from the curse of dimensionality, almost all values for all cluster solutions will differ in very little relative terms and their calculation will also be numerically sensitive.

**Table 1:** Some sensible and non-sensible clustering solutions and their evaluation values based on the existing clustering evaluation metrics

|  | Clustering Techniques | F-measure (higher the better) | Purity (higher the better) | Silhouette Coefficient (higher the better) | XB Index (lower the better) | SSE (lower the better) | DB Index (lower the better) |
|---|---|---|---|---|---|---|---|
| Non-sensible Clustering | GenClust | 0.99 | **0.99** | 0.50 | **0.25** | **65.68** | 0.78 |
|  | HeMI | 0.83 | 0.71 | **0.89** | 0.27 | 2441.59 | **0.13** |
|  | GAGR | 0.99 | 0.98 | 0.13 | 1.03 | 345.66 | 1.55 |
| Sensible Clustering | Example of a reasonable clustering | 0.99 | 0.98 | 0.74 | 0.26 | 1949.67 | 0.33 |

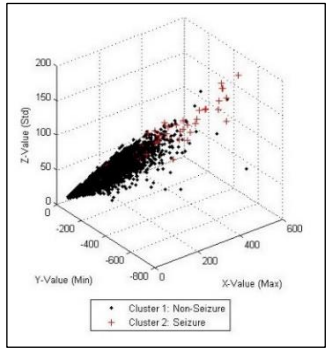

**Fig.1.** The three-dimensional CHB-MIT Scalp EEG (chb01-03) data set

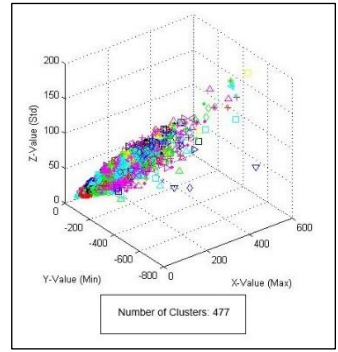

**Fig.2.** Clustering result of GenClust on the CHB-MIT Scalp EEG (chb01-03) data set

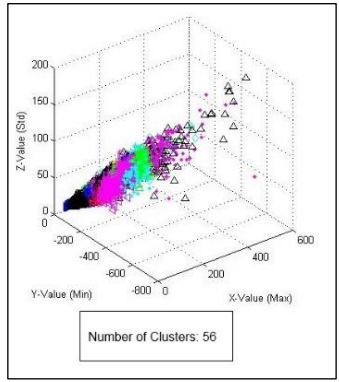

**Fig. 3.** Clustering result of GAGR on the CHB-MIT Scalp EEG (chb01-03) data set

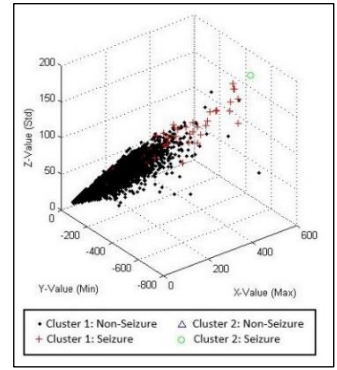

**Fig. 4.** Clustering result of HeMI on the CHB-MIT Scalp EEG (chb01-03) data set

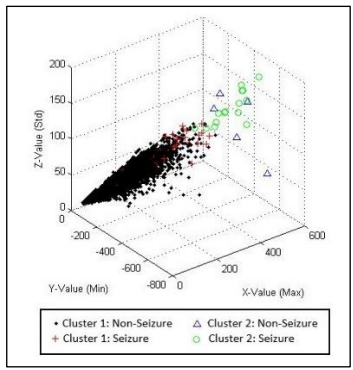

**Fig. 5.** An example of a reasonable clustering result on the CHB-MIT Scalp EEG (chb01-03) data set

Therefore, we realize that an alternative evaluation technique is required for genetically based clustering algorithms. Our approach is inspired in principles where cluster results are a succinct representation of the data (note that even K-means has been identified with vector quantization and error correction). If a clustering results is of high quality (and reasonable), it must produce as well separated cluster that when considered as classes. Well-separated clusters should be easy to represent. The margins among clusters must be sparse. Our Tree Index algorithm aims at describing the structural information of the clustering rather than the quantitative format of cluster-quality indexes. Our Tree Index is finding margins amongst clusters for easy learning without the complications of Minimum Description Length.

**Table 2:** Cluster results of some sensible and non-sensible clustering solutions based on Tree Index

| Clustering Techniques | Tree Index (lower the better) |
|---|---|
| GenClust | 5.36 |
| HeMI | ∞ |
| GAGR | 15.36 |
| Sensible Clustering | **0.14** |

Therefore, we propose a new structural cluster-quality index based on learning decision trees. We will show this Tree Index method can distinguish reasonable and non-sensible clustering solutions. Table 2 shows the clustering results of reasonable and non-sensible clustering solutions based on Tree Index. This is evidence that the proposed evaluation technique is able to identify sensible and non-sensible clustering solutions. It produces a good evaluation value for the sensible clustering solution (shown in Fig. 5) and a bad evaluation value for the non-sensible clustering solutions.

The steps of the proposed clustering evaluation technique are as follows.

*Step 1.* The proposed clustering evaluation technique first labels a data set based on the clustering result that it wants to evaluate. For example: if a clustering technique generates a clustering result with three clusters then Tree Index labels the data set considering the three clusters as three different class values.

*Step 2.* It then builds a decision tree on the labelled data set to classify the records based on their labels. It can use any existing decision tree algorithm. In this study we have used C4.5 [11, 12].

*Step 3.* Tree Index then finds the entropy [1] of each leaf of the tree. The entropy is a well-known evaluation technique that measures the level of uncertainty in a distribution.

*Step 4.* It then finds the depth of each leaf of the tree. Typically, a tree having a lower depth represents a higher agreement between the class labels (and corresponding records of a data set) than a tree with a higher depth.

*Step 5.* It then computes the evaluation value ($M$) as follows.

$$M = \frac{\sum_{i=1}^{l} E_i \times k_i}{|c|} \quad \begin{cases} k_i = d_i \text{ (depth), if } d_i > 0 \\ k_i = \infty, \text{ if } d_i = 0 \end{cases} \quad \text{Eq. (9)}$$

where, $E_i$ is the entropy of the $i^{th}$ leaf, $|c|$ is the number of possible class values, which is the same as the number of clusters, $d_i$ is the depth of the $i^{th}$ leaf. The value of $k_i$ is $d_i$ when the value of $d_i$ is greater than 0. The value of $k_i$ is $\infty$ when the value of $d_i$ is 0. The depth $d_i = 0$ means that the tree has a single leaf with depth zero; that is the root node itself is the only leaf. It means that a tree has not been built indicating that there is no strong pattern in the data set.

This can happen when the records are labelled incorrectly meaning that the clustering results are of poor quality. On the other hand, a good clustering will result in a good labelling of records which will then build a shallow tree with homogeneous leaves (zero entropy). This will obtain a very low $M$ value in Eq. 9.

## 3. RESULTS AND DISCUSSION

### A. The Data Sets and Parameters used in the Experiments

We empirically evaluate our proposed evaluation technique Tree Index in six existing techniques namely K-means [15], K-means ++ [16], AGCUK [3], GAGR [4], GenClust [2] and HeMI [10] on a brain data set (CHB-MIT data set) [13, 14] and Liver Disorder (LD) data set which is available from the UCI machine learning repository [24]. For the experimentation of AGCUK [3], GAGR [4] GenClust [2], HeMI [10] and HeMI++ [18, 23] the population size is set to be 20 and the number of iterations/generations is set to be 50.

In the experiments, the number of iterations of K-means, K-means++, and the number of iterations of K-means in GenClust set to be 50. The number of cluster $k$ in GAGR, K-means and K-means++ is generated randomly in the range between 2 to $\sqrt{n}$, where $n$ is the number of records in a data set. The threshold value for K-means is define as 0.005.

For Tree Index, we need to build a decision tree from a data set where records are labelled on the clustering result that is being evaluated. While building the decision tree we need to assign a minimum number of records for each leaf. In this study we assign 1% of records of a data set, as long as it stays within the range between 2 to 15. If 1% of records is less than 2 then we assign 2, and if 1% of records is more than 15 then we assign 15. We run each technique 20 times on each dataset and we take the average result.

### B. Brain Data Set Pre-processing

For the experimentation, we prepare the CHB-MIT Scalp EEG data set [13, 14] which contains EEG recordings of 22 epileptic patients from different age groups. The EEG signals of the data set were sampled at 256 samples per second with 16-bit resolution.

In the data set most of the cases 23 channels were used, only in some cases 24 or 26 channel were used. For each channel, we divide the data in epochs of 10 seconds. We then calculate the Maximum (Max), Minimum (Min), Mean, Standard deviation (Std), Kurtosis, Skewness, Entropy, Line length and Energy for each epoch. Hence, from each channel of one-hour data we get 360 records containing nine attributes: Max, Min, Mean, Std, Kurtosis, Skewness, Entropy, Line Length and Energy.

In this paper, we prepare one-hour data of one patient (chb01_03) who is an 11 years old girl. This data set has the recordings of 23 channels. Hence, from all 23 channels altogether, we get 360*23=8280 records. In this data set the patient experienced a seizure for around 40 seconds (from the 2996[th] second to 3036[th] second). During this period, we get 5 records. These records are considered as seizure records and all other records are considered as non-seizure records.

Therefore, from the chb_01_03 data set altogether we get 23*5= 115 seizure records and 8165 non-seizure records.

### C. Evaluation of Tree Index on the MIT-chb01_03 data set

In this section, we empirically compare HeMI++ [18, 23] with K-means [15], K-means ++ [16], AGCUK [3], GAGR [4], GenClust [2] and HeMI [10] on a brain data set (MIT-chb01_03) through visual analysis of clustering results. We also compare all the techniques based on Tree Index in order to validate the correctness of Tree Index evaluation. In this section, we use three attributes (Max, Min and Std) of the data set in order to plot the records so that we can see the records and their orientations. Such plots also help us to see clustering results and their appropriateness.

Fig.6 shows the clustering result of HeMI on the CHB-MIT Scalp EEG (chb01-03) data set. HeMI generates two clusters but one cluster contains only one record and all other records belong to the other cluster. Clearly, this does not appear to be a sensible clustering. From Table 4 we can see that according to our proposed cluster evaluation technique HeMI receives a poor evaluation result which is ∞. Therefore, the evaluation made by our proposed Tree Index matches with the manual evaluation (the visual analysis of the plotted records).

Fig. 7 shows the clustering result of AGCUK where it generates two clusters: seizure and non-seizure. Mainly, the non-seizure records appear in Cluster 1 and a mixture of seizure and non-seizure records are found in Cluster 2. Cluster 1 has 2836 non-seizure records (dots in Fig.7) and 0 seizure records (plus signs in Fig.7), while Cluster 2 has 5389 non-seizure records (triangles in Fig. 7) and 55 seizure records (circles in Fig.7). We can clearly see that while the clustering result is more sensible than the clustering result of HeMI (see Fig. 6), it is still not a good clustering result. Our proposed cluster evaluation technique also identifies this. We can see in Table 3 that AGCUK is better than HeMI. This again re-confirms the effectiveness of our proposed evaluation technique.

Fig. 8, Fig. 9, Fig. 10 and Fig. 11 show the clustering results of GAGR, GenClust, K-means and K-means++ where GAGR, GenClust, K-means and K-means++ produce 56, 477, 28 and 13 clusters, respectively. Considering that the data set has only two types of records: Seizure and Non-seizure these clustering results with so many clusters also do not seem appropriate. This is also identified by our proposed evaluation technique as shown in Table 3.

**Table 3:** Clustering results of HeMI++ and other techniques based on Tree Index on MIT-chb01_03 data set

| Clustering Techniques | Tree Index (lower the better) |
|---|---|
| HeMI++ | **0.55** |
| HeMI | ∞ |
| GenClust | 5.27 |
| GAGR | 19.89 |
| AGCUK | 18.19 |
| K-means | 27.41 |
| K-means++ | 31.01 |

As we can see in Fig. 12, HeMI++ [18, 23] produces a sensible clustering solution as it matches with the original orientation of records in the data set (see Fig.5). It produces two clusters: Cluster 1 and Cluster 2. Cluster 1 contains 8219 non-seizure records and 38 seizure records, while Cluster 2 contains 6 non-seizure records and 17 seizure records. As a result, HeMI++ [18, 23] also achieves a good evaluation value based on our proposed evaluation technique Tree Index as shown in Table 3. This re-confirms that Tree Index produces better evaluation value for better clustering solutions.

### D. Evaluation of Tree Index on the LD data set

In order to further evaluation of Tree Index, in this section, we empirically compare the clustering results of all the techniques on the LD data set based on Tree Index. We also graphically visualize the clustering results in order to validate the correctness of Tree Index evaluation. In this section, we use three attributes (mcv mean corpuscular volume, alkphos alkaline phosphatase and sgpt alamine aminotransferase) of the data set in order to plot the records so that we can see the records and their orientations. Fig. 13, Fig. 14, Fig. 15, Fig. 16, Fig. 17, Fig. 18, and Fig. 19 show the clustering results of HeMI, AGUCK, GAGR, GenClust, K-means, K-means++ and HeMI++ [18, 23], respectively. Fig. 20 shows the original structure of the LD data set.

As we can see in Fig. 19, HeMI++ produces a sensible clustering solution as it quite matches with the original orientation of records in the data set. Also, HeMI++ achieves a good evaluation value based on Tree Index evaluation as shown in Table 4. However, it achieves bad evaluation values based on Purity, Silhouette Coefficient, XB Index and DB Index as shown in Table 4.

Fig. 13 shows that HeMI produces non sensible clustering results. It produces two clusters where one cluster contains one record and other clusters contains all the remaining records. It also achieves good evaluation values based on F-measure, Silhouette Coefficient, XB Index and DB Index as shown in Table 4. Similar to HeMI, AGCUK also produces a non-sensible clustering solution (see Fig. 14) and achieves good evaluation values based on F-measure, Silhouette

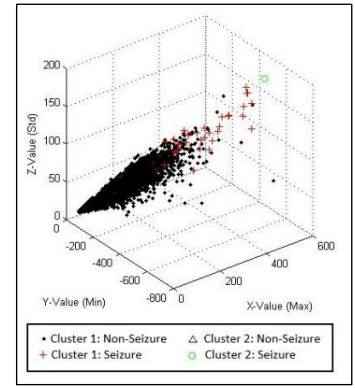

**Fig. 6.** Clustering result of HeMI on the CHB-MIT Scalp EEG (chb01-03

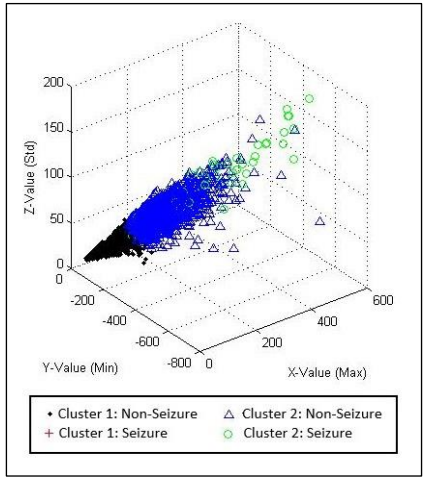

**Fig. 7.** Clustering result of AGCUK on the CHB-MIT Scalp EEG (chb01-03) data set

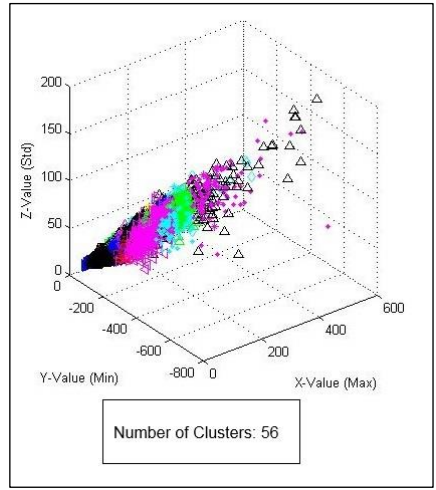

**Fig. 8.** Clustering result of GAGR on the CHB-MIT Scalp EEG (chb01-03) data set

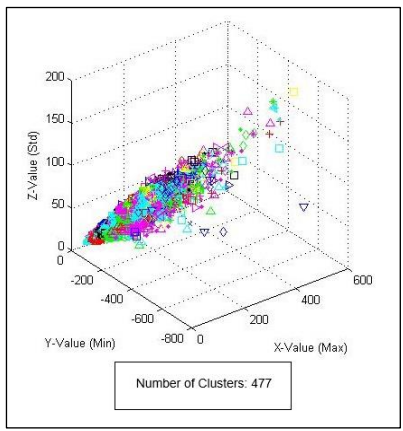

**Fig. 9.** Clustering result of GenClust on the CHB-MIT Scalp EEG (chb01-03) data set

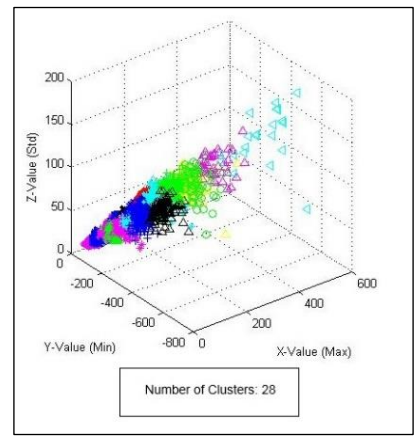

**Fig. 10.** Clustering result of K-means on the CHB-MIT Scalp EEG (chb01-03) data set

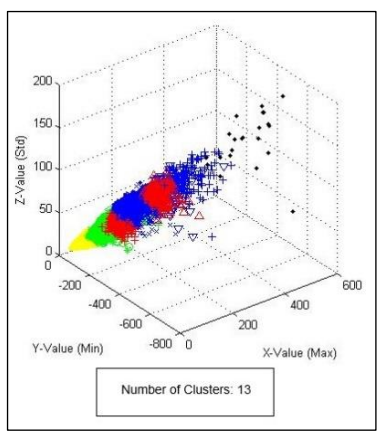

**Fig. 11.** Clustering result of K-means++ on the CHB-MIT Scalp EEG (chb01-03) data set

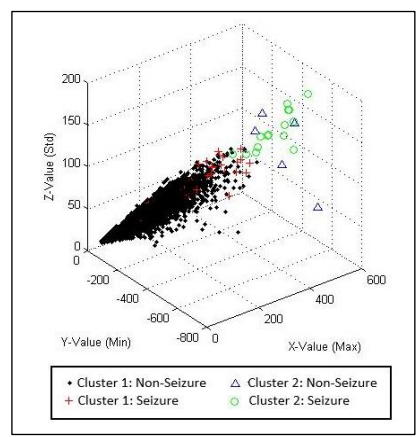

**Fig. 12.** Clustering result of our proposed technique, HeMI++ on the CHB-MIT Scalp EEG (chb01-03) data set

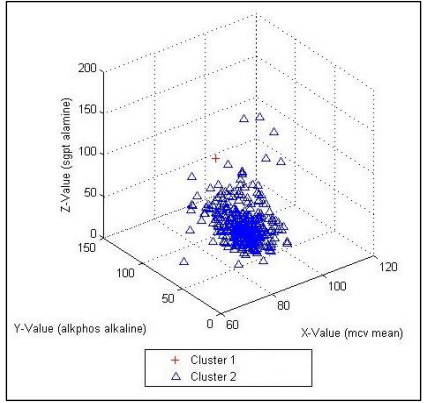

**Fig. 13.** Clustering result of HeMI on the LD data set

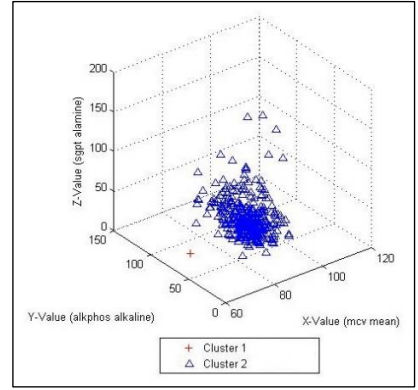

**Fig. 14.** Clustering result of AGCUK on the LD data set

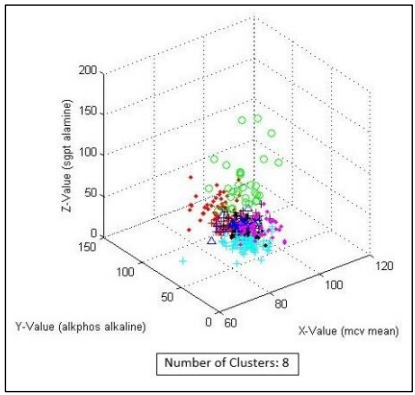

**Fig. 15.** Clustering result of GAGR on the LD data set

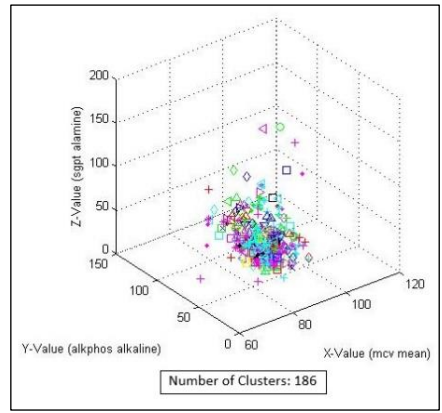

**Fig. 16.** Clustering result of GenClust on the LD data set

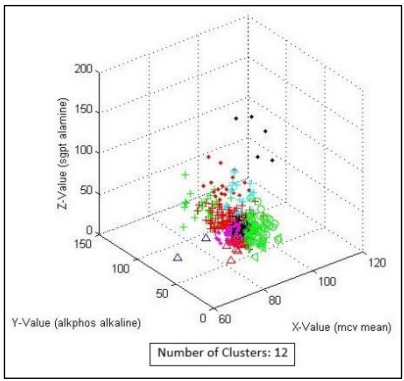

**Fig. 17.** Clustering result of K-means on the LD data set

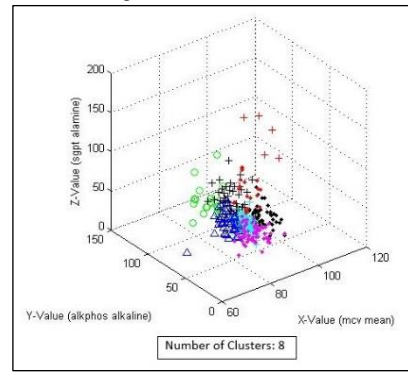

**Fig. 18.** Clustering result of K-means++ on the LD data set

**Table 4:** Clustering results of the techniques on the Liver Disorder (LD) data set based on Tree Index and other evaluation techniques

| Techniques | F-measure (higher the better) | Purity (higher the better) | Silhouette Coefficient (higher the better) | XB Index (lower the better) | SSE (lower the better) | DB Index (lower the better) | Tree Index (lower the better) |
|---|---|---|---|---|---|---|---|
| GenClust | **0.82** | **0.82** | 0.67 | 0.19 | **2.64** | 0.50 | 3.53 |
| HeMI | 0.73 | 0.57 | 0.73 | **0.05** | 100.96 | 0.35 | ∞ |
| GAGR | 0.63 | 0.60 | 0.09 | 1.04 | 59.56 | 1.98 | 5.62 |
| HeMI++ | 0.73 | 0.57 | 0.43 | 0.63 | 95.85 | 0.96 | **0.39** |
| K-Means | 0.60 | 0.60 | 0.21 | 0.34 | 16.60 | 1.10 | 6.21 |
| K-Means++ | 0.65 | 0.59 | 0.12 | 0.23 | 57.27 | 1.13 | 9.62 |
| AGCUK | 0.73 | 0.57 | **0.75** | 0.20 | 33.73 | **0.36** | ∞ |

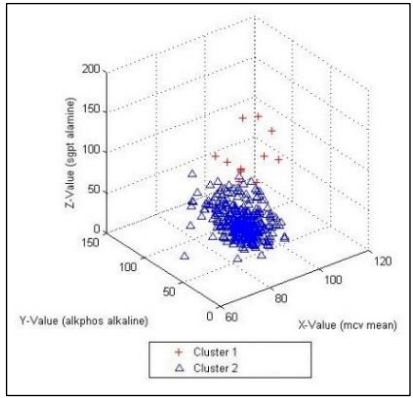

**Fig. 19.** Clustering result of HeMI++ on the LD data set

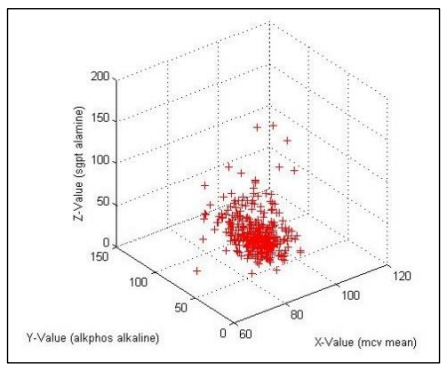

**Fig.20.** The three-dimensional LD data set

Coefficient, XB Index and DB Index as shown in Table 4. However, Tree Index produces bad evaluation values both for HeMI and AGCUK as shown in Table 4. Similarly, Tree Index produces bad evaluation values for other non-sensible clustering results produced by GAGAR, GenClust, K-means and K-means ++ (see column 8 of Table 4). This again re-confirms that Tree Index produces good evaluation value for good clustering solutions and bad evaluation value for bad clustering results.

## CONCLUSION

In this paper we propose a novel cluster evaluation technique called Tree Index. Tree Index first labels the records based on the clustering results that it wants to evaluate. It then builds a decision tree from the data set with the labels. The basic idea here is that if the labelling is good (i.e. sensible) then the produced tree is simpler (shallower) and is likely to classify the training record accurately. Based on this basic concept, our Tree Index computes an evaluation value of a clustering solution. Different clustering solutions can be compared based on their Tree Index values.

**References**

[1] P.-N. Tan, M. Steinbach, V. Kumar, Introduction to Data Mining, first ed., Pearson Addison Wessley, 2005.

[2] M. A. Rahman, & M. Z. Islam, A hybrid clustering technique combining a novel genetic algorithm with K-Means, Knowledge-Based Systems. 71 (2014) 345-365.

[3] Y. Liu, X. Wu, Y. Shen, Automatic clustering using genetic algorithms, Applied Mathematics and Computation. 218 (2011) 267-1279.

[4] D. Chang, X. Zhang, C. Zheng, A genetic algorithm with gene rearrangement for K-means clustering, Pattern Recognition. 42 (2009) 1210-1222.

[5] K.T. Chuang, & M.S. Chen, Clustering Categorical Data by Utilizing the Correlated-Force Ensemble. Paper presented at the 4th SIAM International Conference on Data Mining (SDM 04), Lake Buena Vista, Florida, 2004.

[6] R. Kashef, & M. S. Kamel, Enhanced bisecting -means clustering using intermediate cooperation, Pattern Recognition. 42(2009) 2557-2569.

[7] D.L. Davies, D.W. Bouldin, A cluster separation measure, IEEE Transactions on Pattern Analysis and Machine Intelligence. 1 (1997) 224-227.

[8] L. E. A. Blas, S. S. Sanz, S. J. Fernández, L. C. Calvo, J. D. Ser, J. A. P. Figueras, A new grouping genetic algorithm for clustering problems, Expert Systems with Applications. 39 (2012) 9695- 9703.

[9] V Estivill-Castro. SIGKDD explorations : newsletter of the Special Interest Group on Knowledge Discovery and Data Mining, 2002, Vol.4(1), p.65.

[10] A.H. Beg, & M.Z. Islam, Genetic algorithm with healthy population and multiple streams sharing infrmaton for Clustering, Knowledge-Based Systems, 114 (2016) 61-78.

[11] J. R. Quinlan, C4.5: Programs for Machine Learning, Morgan Kaufmann Publishers, San Mateo, U.S.A., 1993.

[12] J. R. Quinlan, Improved use of continuous attributes in c4.5, Journal of Artificial Intelligence Research 4 (1996) 77–90

[13] A. L. Goldberger, L. A. N. Amaral, L. Glass, J. M. Hausdorff, P. Ivanov, R. G. Mark, J. E. Mietus, G. B. Moody, C.K. Peng, H. E. Stanley, PhysioBank, PhysioToolkit, PhysioNet: Components of a New Research Resource for Complex Physiologic Signals. Circulation 101(23):e215-e220 [Circulation Electronic Pages; http://circ.ahajournals.org/cgi/content/full/101/23/e215]; 2000 (June 13).

[14] A. Shoeb. *Application of Machine Learning to Epileptic Seizure Onset Detection and Treatment*. PhD Thesis, Massachusetts Institute of Technology, September 2009.

[15] S. P. Lloyd, Least squares quantization in PCM, IEEE Transactions on Information Theory. 28 (1982) 129-13.

[16] D. Arthur, & S. Vassilvitskii, k-means++: The Advantages of Careful Seeding, SODA '07 Proceedings of the eighteenth annual ACM-SIAM symposium on Discrete algorithms, 2007, pp.1027-1035.

[17] V Estivill-Castro. The instance easiness of supervised learning for cluster validity, Pacific-Asia Conference on Knowledge Discovery and Data Mining, pp. 197-208, 2011.

[18] A. H. Beg, M. Z. Islam & V.Castro, HeMI++: A Novel Genetic Algorithm-Based Clustering Technique for Sensible Clustering Solution, In Proc. of the IEEE Congress on Evolutionary Computation (IEEE CEC 2020), (Status: Accepted).

[19] S. Montani, & G. Leonardi, Retrieval and clustering for supporting business process adjustment and analysis, Information Systems. 40(2014) 128-141.

[20] A. Mukhhopadhyay, & U. Maulik, Towards improving fuzzy clustering using support vector machine: Application to gene expression data, Pattern Recognition. 42 (2009) 2744-2763.

[21] M. Girvan, & M.E.J. Newman, Community Structure in Social and Biological Networks, Proceedings of the National Academy of Sciences of the United States of America. 99 (2002) 7821-7826.

[22] G. Stockman, & L.G. Shapiro, Computer Vision. (1st ed.), Prentice-Hall, ISBN 0-13-030796-3, New Jersey, 2001,pp. 279-325.

[23] A. H. Beg, A Novel Genetic Algorithm based Clustering and Tree based validation in Producing and Evaluating Sensible Clusters, PhD thesis in Computer Science, Charles Sturt University, 2017.

[24] UCI Machine Learning Repository. <http://archive.ics.uci.edu/ml/data sets.html/> (accessed 22.06.16).